\documentclass[runningheads]{llncs}
\usepackage[T1]{fontenc}
\usepackage{graphicx}
\usepackage{amsfonts}
\usepackage{amsmath}
\usepackage{multirow}
\DeclareMathOperator{\mlp}{MLP}
\DeclareMathOperator{\transformer}{Transformer}
\DeclareMathOperator{\mean}{Mean}
\DeclareMathOperator{\argtop}{argtop}
\DeclareMathOperator{\maxpool}{Maxpool}
\begin{document}
\title{Joint Top-Down and Bottom-Up Frameworks \\ for 3D Visual Grounding}
%
%
\author{Yang~Liu \and
Daizong~Liu \and
Wei~Hu}
\authorrunning{Liu et al.}
%
\institute{Wangxuan Institute of Computer Technology, Peking University, Beijing, China\\
\email{2018liuyang@pku.edu.cn, dzliu@stu.pku.edu.cn, forhuwei@pku.edu.cn}}
\maketitle              
\begin{abstract}
This paper tackles the challenging task of 3D visual grounding—locating a specific object in a 3D point cloud scene based on text descriptions. Existing methods fall into two categories: top-down and bottom-up methods. Top-down methods rely on a pre-trained 3D detector to generate and select the best bounding box, resulting in time-consuming processes. Bottom-up methods directly regress object bounding boxes with coarse-grained features, producing worse results. To combine their strengths while addressing their limitations, we propose a joint top-down and bottom-up framework, aiming to enhance the performance while improving the efficiency. Specifically, in the first stage, we propose a bottom-up based proposal generation module, which utilizes lightweight neural layers to efficiently regress and cluster several coarse object proposals instead of using a complex 3D detector.
Then, in the second stage, we introduce a top-down based proposal consolidation module, which utilizes graph design to effectively aggregate and propagate the query-related object contexts among the generated proposals for further refinement. By jointly training these two modules, we can avoid the inherent drawbacks of the complex proposals in the top-down framework and the coarse proposals in the bottom-up framework. Experimental results on the ScanRefer benchmark show that our framework is able to achieve the state-of-the-art performance.

\end{abstract}
\begin{figure}[t!]
\centering
\includegraphics[width=0.8\columnwidth]{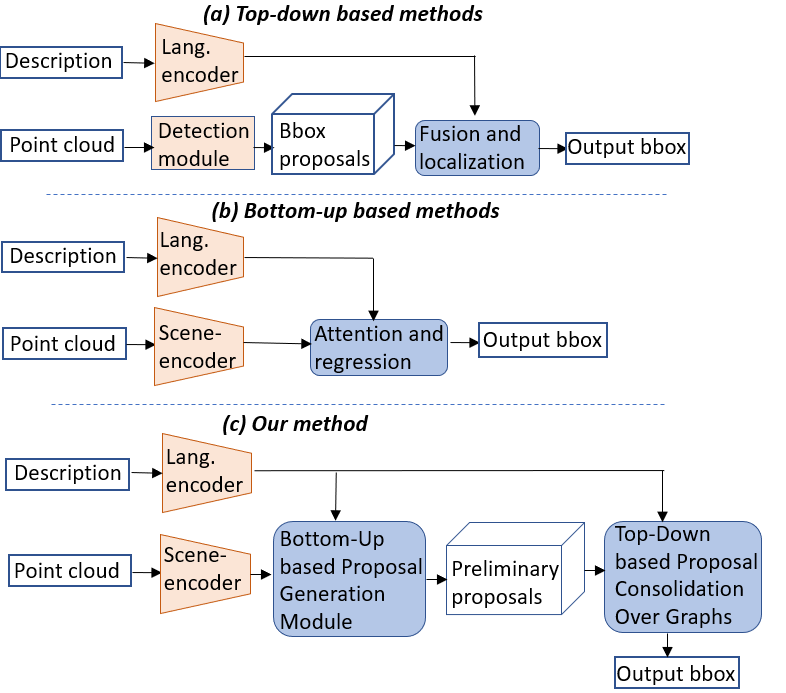}
\vspace{-12pt}
\caption{(a): Typical procedure of top-down based method. (b): Typical procedure of bottom-up based method. (c): Procedure of our proposed method, where we generate initial proposals in an efficient bottom-up manner, and subsequently consolidate the proposals over graphs via an effective top-down approach.}
    \label{fig:frame}
    \vspace{-12pt}
\end{figure} 

\section{Introduction}
The 3D visual grounding (3DVG) \cite{chen2020scanrefer,liu2024survey,huang2023dense} is a fundamental yet important task in 3D understanding, which has recently received increasing attention due to its wide range of applications \cite{liu2022imperceptible,hu2023density,hu2022exploring,liu2023point,tao20233dhacker,liu2024a,liu2023robust,liu2024explicitly}, such as in robotics and AR/VR systems. The goal of this task is to locate the target object in a 3D point cloud scene based on a given free-form query text description. Different from previous mature 2D grounding methods \cite{wang2018learning,wang2019neighbourhood,deng2021transvg,yang2022tubedetr,li2021referring,sadhu2019zero,yang2019fast,yang2019cross,liu2021context,liu2020jointly,liu2020saanet,liu2021video,liu2021f2net,liu2023exploring,liu2022skimming,liu2021spatiotemporal,liu2022learning,liu2022rethinking},  3D visual grounding has two more challenging aspects: Firstly, 3D visual grounding takes sparse, noisy, and information-dense 3D point clouds as input, making it more difficult to obtain visual information. Secondly, the object-to-object and object-to-scene relationships in 3D space are more complex than that in 2D images, further increasing the difficulty of 3D visual grounding task.

Existing methods for the 3D visual grounding (3DVG) task can mainly be grouped into two categories according to their model designs:
(1) Top-down approaches: these methods \cite{chen2020scanrefer,feng2021free,yuan2021instancerefer,huang2021text,yang2021sat,achlioptas2020referit3d,liu2024cross} typically first utilize pre-trained 3D object detection models \cite{qi2019deep,liu2021group,vu2022softgroup,cheng2021back,misra2021end} or segmenter \cite{chen2021hierarchical,jiang2020pointgroup} to generate a large number of candidate proposals in the entire point cloud scene, and then select the one that best matches the semantic meaning of the query text. 
Although they are able to obtain high-quality proposals, they need to enumerate all possible objects to ensure that the generated proposals contain the object required by the text, leading to a large number of redundant proposals. The general procedure of the top-down based methods is illustrated in Figure \ref{fig:frame}(a).
(2) Bottom-up approaches: these methods \cite{luo20223d,liu2021refer,he2021transrefer3d} generally first interact the point set with textual description via early-fusion strategy, then directly predict the target bounding box from the learned query-related point-wise features. 
Compared to the top-down methods,
they do not rely on complex proposal generation and selection, and thus can achieve end-to-end efficient training. The general procedure of the bottom-up based methods is illustrated in Figure \ref{fig:frame}(b).

However, the above two types of methods have their own advantages and disadvantages. 
For the top-down approaches, this category of methods can typically yield high-quality proposals and facilitate the better capture of relationship information among a large number of proposals. However, its disadvantage is that in order to ensure that the generated proposals contain the proposal required by the query text, a large number of proposals need to be generated in abundance, while the vast majority of the generated proposals are not related to the description in the query text, which reduces the efficiency of the method. However, if the number of generated proposals is reduced, there is a higher risk of ignoring the object required by the query. 
For the bottom-up approaches, because they avoid the issue of generating and processing an excessive number of redundant proposals, higher computational efficiency is achievable. At the same time, since the bounding box is directly regressed and not limited to the proposals generated by the pre-trained model, this category of methods can capture smaller objects that may be easily neglected by the pre-trained model. However, they overlook the rich information between the global points as they struggle to model object-level interactions. Therefore, their predicted object proposals obtained are relatively coarse, and there is no additional design for further proposal refinement.

Through the analysis above, we find that the advantages and disadvantages of these two methods actually complement each other. Specifically, bottom-up approaches can generate a few proposals that closely relate to the query text, thereby reducing redundancy caused by top-down approaches and significantly improving computational efficiency. On the other hand, top-down approaches can alleviate the issue of coarse proposals generated by bottom-up approaches by capturing the relationship information among proposals and refining them.

This inspires us to propose a method that integrates the advantages of both approaches while mitigating their limitations. Our proposed method consists of two stages. In the first stage, we develop a \textit{proposal generation module} similar to bottom-up methods. This module aggregates from both 3D point clouds and query texts, enabling us to predict bounding boxes for objects highly relevant to the query text directly based on these features, and extract the corresponding object features. This enables us to use the guidance of the query text information to avoid the inefficiency caused by detecting and analyzing a large number of redundant objects simultaneously, and to identify objects that may be overlooked by pre-trained detectors. In the second stage, we address the issue of rough proposals generated by bottom-up methods by developing a \textit{graph-based proposal consolidation module} inspired by top-down methods. This module further captures object-to-object and object-to-scene relationships and updates the features of objects accordingly. Subsequently, we generate more refined bounding boxes for the objects based on these updated features. We conducted evaluations on the commonly adopted ScanRefer \cite{chen2020scanrefer} datasets, and the experimental results demonstrated that our proposed method achieved the state-of-the-art performance when compared to existing methods.

In summary, the contributions of our work are:
\begin{enumerate}
    \item We conduct an in-depth analysis of the strengths and weaknesses of existing top-down and bottom-up-based methods for 3D visual grounding. Through this analysis, we provide valuable insights into how to leverage their respective advantages while mitigating their limitations effectively.
    \item We propose a novel framework, which first develops a bottom-up based strategy to generate a few object proposals and then devises a top-down based strategy over graphs to consolidate and refine the proposals for final grounding.
    \item Through comprehensive experiments, we demonstrate the effectiveness of our proposed method and shed light on the rationale behind the successful integration of bottom-up and top-down approaches.
\end{enumerate}

\section{Related Work}
\noindent \textbf{Top-down based 3D visual grounding.}
Most approaches for 3DVG are top-down based. For example, ScanRefer \cite{chen2020scanrefer} utilizs VoteNet \cite{qi2019deep} to extract numerous proposals and combine their features with textual features to select the matched proposal.
Subsequently, several top-down approaches emerged. TGNN \cite{huang2021text} generates candidate proposals as graph nodes, leveraging object features and relationships to generate attention heatmaps for sentence expressions. InstanceRefer \cite{yuan2021instancerefer} uses a language model to determine the target object category and identifies instances with the same category in the scene as candidates. The final instance is chosen through a matching process.
SAT \cite{yang2021sat} enhances understanding of 3D scenes by learning alignments between 2D object representations and corresponding objects in 3D scenes
While these methods aim to include desired proposals specified by textual requirements, generating a large number of proposals often leads to inefficiencies. Besides, reducing generated proposals increases the risk of neglecting query-required objects.

\noindent \textbf{Bottom-up based 3D visual grounding.}
To address the challenge of generating numerous irrelevant proposals in top-down methods, bottom-up method 3D-SPS \cite{luo20223d} is proposed to progressively selects key points based on language guidance and directly regresses the bounding box. Similarly, Refer-it-in-RGBD \cite{liu2021refer} first constructs a confidence heat map from the input sentence and voxels, then samples seed points according to the heat map, and regresses the object's bounding box.
However, these bottom-up methods often produce coarse proposals, limiting their ability to exploit complementary information among different bounding boxes for refinement.

\begin{figure*}[t!]
\centering
\includegraphics[width=1.0\columnwidth]{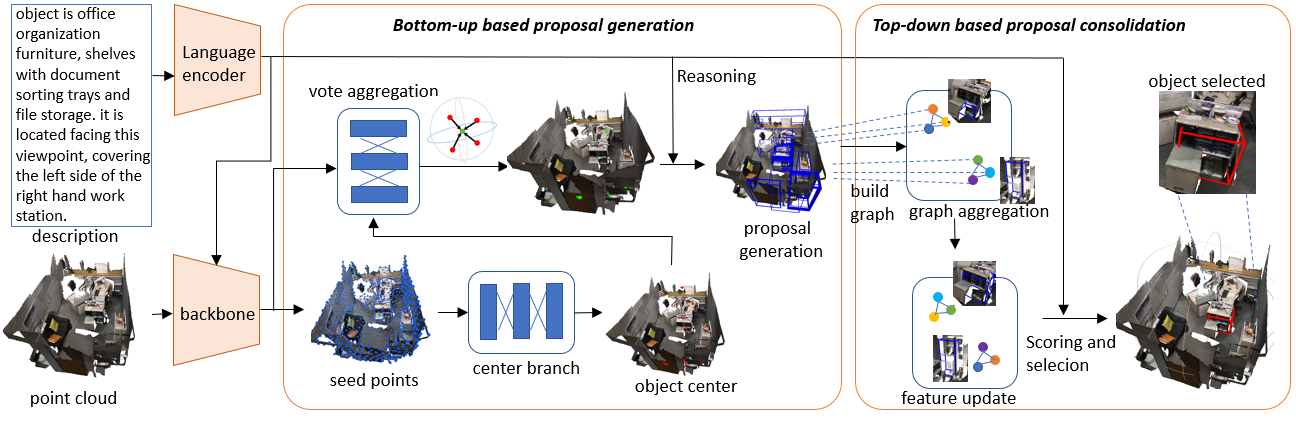}
\caption{The pipeline of our proposed method. Initially, we encode the input 3D point cloud and text with pre-trained encoders. In the bottom-up stage, our module fuses these features for language-guided object proposals. In the top-down stage, our refinement module enhances these proposals by graph-based features, followed by predicting matching scores to select the best-matching bounding box.}
    \label{fig:pipeline}
    \vspace{-10pt}
\end{figure*} 

\section{The Proposed Method}\label{The Proposed Method}
Previous works generally follow a top-down or bottom-up framework, both of which come with inherent limitations within their respective designs.
In this paper, we propose a novel approach that leverages the strengths of both top-down and bottom-up frameworks while mitigating their individual limitations through a unified structure.
In this section, we provide a comprehensive description of our method. We begin by offering an overview of the 3D visual grounding task and our proposed framework. Then, we describe the multi-modal encoders used in our method. After that, we elaborate on our proposed bottom-up based 3D proposal generation module and top-down based 3D proposal consolidation module respectively. At last, we present the training objectives of our method.

\subsection{Overview}\label{Overview}
\noindent \textbf{Notation definition.}
We first define 3D visual grounding task as follows. Given point clouds $\mathbf{P} \in \mathbb{R}^{N \times (3+F)} $ and free-form language query text $\mathbf{D} = \{w_n\}_{n=1}^{W}$ , where $N$ is the number of the points, $F$ is the dimension of the additional features of the point clouds such as colors and normals , and $W$ denotes the number of the input words. Our task is to predict the 3D bounding box of the object that matches the input description.

\noindent \textbf{Overall pipeline.}
As illustrated in Figure \ref{fig:pipeline}, our proposed method
consists of two stages: the bottom-up stage and the top-down stage. Firstly, we use pre-trained encoders to independently encode the 3D point cloud and the query text information. In the bottom-up stage, we feed these two types of features into our proposed bottom-up based proposal generation module for feature fusion and updating, which yields language-guided object proposals. Subsequently, in the top-down stage, these proposals are input into our proposed top-down based proposal consolidation module for further refinement, resulting in improved proposals. Finally, we predict the matching scores between these proposals and the query language, selecting the bounding box that best matches the query language as the final output.

\subsection{Preliminaries}\label{Preliminaries}
\noindent \textbf{3D scene encoder.}
There have been many works \cite{qi2017pointnet,qi2017pointnet++,wang2019dynamic,radosavovic2020designing} on encoding 3D point clouds, and theoretically, they can all be used for encoding the input 3D scene point clouds. For consistency with previous works on 3D visual grounding \cite{chen2020scanrefer,liu2021refer,yang2021sat}, we adopt the same pre-trained PointNet++ \cite{qi2017pointnet++} for encoding the 3D point cloud information. Let $\mathbf{V}\in \mathbb{R}^{M_0 \times (3+C_v)} $ denote the output of PointNet++, where $M_0$ is the number of the seed points obtained by PointNet++ and $C_v$ is the dimension of point features. Each point's feature $V_i$ can be divided into two parts, which represent its 3D coordinates $ \mathbf{x}_i\in \mathbb{R}^{3}$ and other features $\mathbf{f}_i\in \mathbb{R}^{C_v}$.

\noindent \textbf{Description encoder.}
We use a pre-trained CLIP model \cite{radford2021learning} to encode the language information from the query text $\mathbf{D} = \{w_n\}_{n=1}^{W}$. The output of the text encoder is denoted as $\mathbf{L}\in \mathbb{R}^{W \times C_l} $, where $C_l$ here represents the dimension of language features.

\subsection{Bottom-Up based Proposal Generation Module}\label{3D Proposal Generation Module}

To locate the object, previous top-down based methods directly utilize 3D detection models to produce all possible proposals of all objects in the 3D scene, which not only severely rely on the proposal quality but also result in low computational efficiency. Although some recent bottom-up based methods try to directly regress query-related proposals, they still rely on the complex decoding modules.
 
Therefore, we propose a simple yet lightweight bottom-up based proposal generation module that generates candidate proposals guided by the query text. This approach achieves higher efficiency compared to traditional top-down based methods and is simpler than previous bottom-up based approaches.

To achieve this goal, we first utilize a multi-modal transformer-based module \cite{vaswani2017attention} to fuse language and 3D visual information and eliminate points that are irrelevant to the given language query. By leveraging the features generated by the Transformer, we predict the center coordinates of the objects to which each point in the point cloud belongs. Subsequently, based on each center, we employ a vote-aggregation module to combine features of its neighboring points belonging to the same object. Using these aggregated features, we can regress language-guided object proposals.

Specifically, as for the multi-modal transformer-based module, we employ two separate self-attention layers to encode the contexts in 3D point cloud features and query language features, along with a cross-attention layer to encode the correspondences between the two modalities. The self-attention mechanism allows the model to capture relevant relationships and dependencies within each modality, while the cross-attention mechanism helps in aligning the language and visual features, enabling the model to focus on the most relevant information for the task of generating object proposals guided by the language query. The output of the transformer module is denoted as $\mathbf{V}^T \in \mathbb{R}^{M_0 \times C_t}$, and the computation process can be represented as follows:
\begin{equation}
    \mathbf{V}^T = \transformer(\mathbf{V},\mathbf{L}) .
\end{equation}

By examining the attention scores produced by the cross-attention mechanism, we can filter out points that have weak associations with the given query language. Denoting the attention scores as $att_{lang}$, we obtain the filtered set of points $\mathbf{V}^F \in \mathbb{R}^{M_f \times C_t}$ through the following computation: 
\begin{equation}
    \mathbf{V}^F=\mathbf{V}_t[\argtop_k(\mean(att_{lang}),M_f)] .
\end{equation}

Subsequently, for each of the points obtained above, we employ a center predictor based on its features $\mathbf{V}^F$ to predict the object it belongs to.  This process yields the coordinates of the center of the corresponding object for each point, denoted as $\mathbf{c}_f \in \mathbb{R}^{M_f \times 3}$. Here, $\mathbf{c}_f$ represents the three-dimensional coordinates of $M_f$ individual center points.. The center predictor can be implemented using a Multi-Layer Perceptron(MLP). Then, to obtain object candidate regions from the points, we first employ farthest point sampling based on these center coordinates to obtain several candidate proposal centers $\{c_i\}_{i=1}^K$, where $K$ is the number of candidates. Since points belonging to the same object have closely related center coordinates, farthest point sampling helps to select those points that belong to different objects. After obtaining the selected points using farthest point sampling, we perform max pooling on the points within a certain distance from each selected center. Let the set of points within a certain distance $r$ from point $c_i$ be denoted as $\{c_{ij}\}_{j=1}^n$ and their features denoted as $\{\mathbf{V}_{ij}^F\}_{j=1}^n$, the feature $\mathbf{V}_{i}^G$ of point $c_i$ can be computed as:
\begin{equation}
    \mathbf{V}_{i}^G=\maxpool(\{\mlp(\mathbf{V}_{ij}^F)\}_{j=1}^n).
\end{equation}
The purpose of this max-pooling operation is to aggregate information from points belonging to the same object.

The aggregated features$\{\mathbf{V}_{i}^G\}_{i=1}^K$ are then fed into a Proposal Predictor, which employs an MLP to regress preliminary proposal results. The Proposal Predictor predicts the bounding box center $\hat{\mathbf{c}_0}$ and bounding box size $\hat{\mathbf{r}_0}$ for each point that defines the proposed regions corresponding to different objects in the 3D scene:
\begin{equation}
    [\hat{\mathbf{c}_0},\hat{\mathbf{r}_0}]=\mlp(\mathbf{V}^G).
\end{equation}

\subsection{Top-Down based Proposal Consolidation Module}\label{3D Proposal Consolidation Module}
Since the object proposals are not always accurate, \textit{i.e.}, it may contain only a part of the object or include surrounding objects within the bounding box, it is crucial to refine them.
However, previous bottom-up based methods directly output the regressed proposals as the final result, easily resulting in inaccurate localization. 
Although top-down based methods try to correlate all proposals for selecting the best one, they still haven't refined the proposals. 
To address these issues, we propose to construct a graph structure to learn the correlations between the proposals. Different from previous methods, we additionally develop a weighted edge for solely correlating the relevant proposals. Moreover, we further devise a novel proposal consolidation strategy to enrich and refine the information in each proposal based on the contexts from its relevant proposals belonging to the same object.

Specifically, we aim to refine each proposal using a graph-based method that leverages information from its neighboring and relevant proposals. To achieve this, we start by constructing a fully connected graph among the generated proposals. However, we recognize that the correlations between different nodes (objects) in the graph can vary significantly. Objects that are close in spatial location and share similar categories often have stronger relationships. To make the most of these correlations between nodes while reducing interference from unrelated nodes, we define the edge weights $W_{uv}$ between proposals as follows:
\begin{equation}\label{graph}
    \mathbf{W}_{v u}=\left\{\begin{array}{cc}
\alpha \cos \left(\mathbf{V}^G_v, \mathbf{V}^G_u\right)+\beta \operatorname{IoU}\left(\mathbf{c}_v, \mathbf{c}_u\right), & v \neq u \\
1, & v=u.
\end{array}\right.
\end{equation}

Here, $\alpha$ and $\beta$ are hyper-parameters that quantify the weights of the two terms. The first term represents the semantic correlation strength between two proposals, measured by their cosine similarity of features. The second term quantifies the spatial correlation between the two proposals based on the Intersection over Union (IoU) value of their preliminary bounding boxes. The bounding box can be computed using the center $\hat{\mathbf{c}_0}$ and size $\hat{\mathbf{r}_0}$.

Then, we employ the edge weights to update the features of the proposals $\mathbf{V}^A$ through a weighted summation process:
\begin{equation}
    \mathbf{V}^A_v = \sum_u{\mathbf{W}_{v u}\mathbf{V}^G_u}.
\end{equation}
Through this step, we refine the representation of the proposals, taking into account their semantic and spatial associations with other relevant proposals in the scene.

\subsection{Training Objectives}\label{Training}

Based on the enriched features of the proposals $\mathbf{V}^A$, we utilize a proposal predictor to refine the bounding box information $[\hat{\mathbf{c}},\hat{\mathbf{r}}]$ for each proposal along with the matching score $\hat{s}$ between the proposal and the query text.  We then select the proposal with the highest matching score as the final result.

To ensure the model achieves satisfactory results, we adopt a combination of multiple loss functions to supervise the entire pipeline.

During the preliminary proposal generation process, we utilize a center loss $\mathcal{L}_{\text{center}}$ to supervise the center predictor and ensure the correct prediction of object center coordinates. The center loss is computed by calculating the L1 loss between the predicted center coordinates $\mathbf{c}_f$ and the ground truth center coordinates $\mathbf{c}_{gt}$ corresponding to that point. 
\begin{equation}
    \mathcal{L}_{\text{center}}=\Vert \mathbf{c}_f-\mathbf{c}_{gt} \Vert_1.
\end{equation}
To obtain the ground truth center for a point, we leverage the dataset's object centers. For each point, we calculate its distance to all the ground truth object centers in the dataset. The point is then assigned to the object with the closest center, making it the ground truth center for that point.

In the final selection stage, we employ a reference loss $\mathcal{L}_{\text{ref}}$ to supervise the selection of the best matching proposal for the query text. For each proposal, with its matching score $s_i$ and corresponding bounding box $[\hat{\mathbf{c}},\hat{\mathbf{r}}]$, we determine its corresponding label using the following approach. First, we calculate the IoU between the predicted bounding box and the ground truth bounding box. Then, we set the label of the proposal $y_i$ to 1 if it has the maximum IoU value among proposals and the IoU value is greater than a specified threshold. For all other proposals, the label is set to 0. The reference loss $\mathcal{L}_{\text{ref}}$ is computed by calculating the cross-entropy between the predicted matching score $s$ and the corresponding label $y$. The formula for the reference loss is as follows:
\begin{equation}
    \mathcal{L}_{\text{ref}}=-\sum_i(y_i\log(s_i)+(1-y_i)\log(1-s_i)).
\end{equation}

Additionally, similar to some previous works \cite{chen2020scanrefer,qi2019deep}, we incorporate an object detection loss $\mathcal{L}_{\text{det}}$ to supervise the object detection process. This loss function helps the model accurately localize and identifies objects within the scene.
Moreover, we utilize a Language to Object classification loss $\mathcal{L}_{\text{lang}}$ to aid in language understanding.

\noindent \textbf{Overall loss.}
By combining these different loss functions, our proposed approach aims to achieve improved performance in 3D visual grounding task. As a summary, the total loss used during the training process can be represented as:
\begin{equation}
    \mathcal{L}=\lambda_1 \mathcal{L}_{\text{center}}+ \lambda_2 \mathcal{L}_{\text{ref}}+\lambda_3 \mathcal{L}_{\text{det}}+\lambda_4 \mathcal{L}_{\text{lang}},
\end{equation}
where $\lambda_1$, $\lambda_2$, $\lambda_3$ and $\lambda_4$ are the weights assigned to different types of losses.

\section{Experiments}

\subsection{Datasets and Evaluation Metric}
To validate the effectiveness of our method and compare with previous works, we conducted experiments on the commonly used ScanRefer dataset \cite{chen2020scanrefer} and two Referit3D dataset Nr3D and Sr3D \cite{achlioptas2020referit3d}. ScanRefer is designed for the 3D visual grounding task. It contains a total of 51,583 textual descriptions corresponding to the objects provided in 806 scanned scenes from the ScanNet dataset. On average, each scene contains 13.81 objects, and each object is associated with 4.67 textual descriptions in the ScanRefer dataset. 

To evaluate the performance, we utilize the metric Acc@kIoU, where 'k' represents the threshold for the IoU between the predicted bounding box and the ground truth. Following previous works, we set 'k' to 0.25 and 0.5 for our experiments. 

Nr3D and Sr3D \cite{achlioptas2020referit3d} provides 41.5K and 83.6K textual descriptions for scenes in ScanNet, respectively. We evaluate the effectiveness of our method on NR3D using the same metrics.

\subsection{Implementation Details}
During the training process, we employed 4 NVIDIA RTX3090 GPUs, with a batch size of 6 on each of the 4 GPU, resulting in a total effective batch size of 24. The training process was conducted for 32 epochs. We utilized the AdamW optimizer \cite{loshchilov2018fixing} with an initial learning rate of 0.001 for optimization. The pretrained language model is frozen. During the k-th epoch, the learning rate was calculated using the following formula:
\begin{equation}
    lr(k) = 0.5\times \left(1+\cos{\frac{(k-1)\pi}{32}}\right)\times0.001.
\end{equation}
During the encoding stage, we employed pre-trained PointNet++ and CLIP models.  The input information includes point cloud coordinates, normal vectors, color vectors and 2D multiview features. The number $M_0$ of the output points of pointnet++ is 2048. In the bottom-up proposal generation module, the number of points $M_f$ filtered based on attention coefficients is set to 512. Afterward, the number of points $K$ selected using the FPS is set to 128. In the top-down based proposal consolidation module, the coefficients $\alpha$ and $\beta$ in Equation \ref{graph} are set to 0.7 and 0.3, respectively. As for the loss function, the coefficients $\lambda_1$, $\lambda_2$, $\lambda_3$ and $\lambda_4$ are set to 5, 0.1, 5, and 0.1, respectively.

\begin{table*}[t!]
\centering
\small
\caption{Comparison on ScanRefer dataset.}
\vspace{-8pt}
\begin{tabular}{l|cc|cc|cc}
\hline
\multicolumn{1}{c|}{\multirow{2}{*}{Method}} & \multicolumn{2}{c|}{unique}                                 & \multicolumn{2}{c|}{multiple}                               & \multicolumn{2}{c}{overall}                                \\
\multicolumn{1}{c|}{}                        & \multicolumn{1}{l}{Acc@0.25} & \multicolumn{1}{l|}{Acc@0.5} & \multicolumn{1}{l}{Acc@0.25} & \multicolumn{1}{l|}{Acc@0.5} & \multicolumn{1}{l}{Acc@0.25} & \multicolumn{1}{l}{Acc@0.5} \\ \hline \hline
ScanRefer \cite{chen2020scanrefer}                                   & 76.33                        & 53.51                        & 32.73                        & 21.11                        & 41.19                        & 27.40                       \\
InstanceRefer    \cite{yuan2021instancerefer}                            & 75.72                        & 64.66                        & 29.41                        & 22.99                        & 38.40                        & 31.08                       \\
SAT      \cite{yang2021sat}                                    & 73.21                        & 50.83                        & 37.64                        & 25.16                        & 44.54                        & 30.14                       \\
3DVG-Transformer       \cite{zhao20213dvg}                      & 81.93                        & 60.64                        & 39.30                        & 28.42                        & 47.57                        & 34.67                       \\
3D-SPS (reported)    \cite{luo20223d}                                   & \textbf{84.12}                        & 66.72                        & 40.32                        & 29.82                        & 48.82                        & 36.98                       \\
3D-SPS (Re-imple)    \cite{luo20223d}                                   & 82.82                        & 64.77                        & 39.58                        & 29.11                        & 47.97                        & 36.03                       \\
\textbf{Ours}                                & 82.60                       & \textbf{66.83}                        & \textbf{40.96}                        & \textbf{30.81}                        & \textbf{49.04}                     & \textbf{37.80}                      \\ \hline
\end{tabular}
\label{scanrefer}
\end{table*}

\subsection{Comparison with SOTA}
We compared our experimental results on the ScanRefer dataset with those of previous methods, as shown in Table \ref{scanrefer}. In our evaluation, we measured the proportion of predicted bounding boxes with an IoU greater than 0.25 and 0.5 concerning the ground truth bounding boxes. All methods in the table utilize 3D point cloud features combined with 2D multiview features as inputs.The methods we compare with encompass both top-down based approaches, such as ScanRefer \cite{chen2020scanrefer}, InstanceRefer \cite{yuan2021instancerefer}, SAT \cite{yang2021sat}, and 3DVG-Transformer \cite{zhao20213dvg}, as well as bottom-up based approaches, such as 3D-SPS \cite{luo20223d}. To provide a comprehensive analysis, we followed the example of previous works \cite{chen2020scanrefer} and presented the results separately for the "unique" and "multiple" subsets. The "unique" subset refers to scenes where there is only one object of the same category as the target object, while the "multiple" subset includes scenes with multiple objects of the same category. In the table, "reported" represents the results reported in the paper, while "our implementation" refers to the results of our reproduced experiments.

From Table \ref{scanrefer}, our method has achieved the best results in five out of the six metrics.  Particularly, in the two primary metrics, "overall-Acc@0.25" and "overall-Acc@0.5", our method has demonstrated improvements of 0.56 and 1.61 percentage points respectively compared to the previously leading approach. The improvements in accuracy over the previous state-of-the-art method (our implementation) demonstrate the effectiveness of our proposed technique. 

To further validate the effectiveness of our proposed method, we conducted experiments on two datasets from ReferIt3D \cite{achlioptas2020referit3d}, namely NR3D and SR3D. We adopted an experimental setup similar to ScanRefer \cite{chen2020scanrefer}, where we directly predict the bounding box of the object described in the text, without relying on the provided bounding box information. We used a similar evaluation metric, Acc@0.25, for assessment. The experimental results are presented in Table \ref{referit3d}. The results for other methods in the table were also obtained using the same experimental setup. From the Table \ref{referit3d}, it can be observed that our method outperforms all others on both of these datasets.

\begin{table}[t!]
\caption{Comparison on Referit3D dataset.}
\vspace{-8pt}
\centering
\setlength{\tabcolsep}{3.0mm}{
\begin{tabular}{l|c|c}
\hline
\multicolumn{1}{c|}{Method} & SR3D/Acc@0.25 & \multicolumn{1}{l}{NR3D/Acc@0.25} \\ \hline
ReferIt3D                   & 27.7          & 24.0                              \\
InstanceRefer               & 31.5          & 29.9                              \\
LanguageRefer               & 39.5          & 28.6                              \\
SAT                         & 35.4          & 31.7                              \\
\textbf{Ours}               & \textbf{41.4}         & \textbf{32.0}                             \\ \hline
\end{tabular}}
\label{referit3d}
\end{table}

\begin{table}[t!]
\centering
\caption{Ablation study results on ScanRefer dataset using 'overall-Acc@0.25' and 'overall-Acc@0.5' metrics.}
\vspace{-8pt}
\setlength{\tabcolsep}{2.0mm}{
\begin{tabular}{c|c|c|c}
\hline
bottom-up          & \multicolumn{1}{c|}{top-down} & Acc@0.25 & Acc@0.5 \\ \hline
$\times$     & $\times$                & 44.27    & 33.30   \\
$\checkmark$ & $\times$                & 45.30    & 34.32   \\
$\times$     & $\checkmark$           & 45.46    & 33.82   \\
$\checkmark$ & $\checkmark$         &\textbf{49.04}& \textbf{37.80}   \\ \hline
\end{tabular}}

\vspace{-8pt}
\label{ablation}
\end{table}




\subsection{Ablation Study}
\noindent \textbf{Main ablation.} To further demonstrate the effectiveness of our proposed bottom-up based proposal generation module and top-down based proposal consolidation module, we conduct the following experiments on the ScanRefer dataset as shown in Table \ref{ablation}. In the first row, we report the performance achieved using a pre-trained encoder and transformer module, along with a proposal generation method similar to ScanRefer, serving as our baseline for comparison. The second and third rows illustrate the results when our proposed bottom-up module and top-down module are incorporated, respectively. The last row presents the results achieved by the complete model, integrating all components.
From Table \ref{ablation}, it is evident that both of the proposed modules contribute to improving the experimental results. This observation highlights the essential role of both bottom-up and top-down cues in the 3D visual grounding task. Notably, when these two components are combined, they achieve the highest accuracy, demonstrating their complementary roles in addressing the 3D visual grounding task. 

\begin{figure}[t!]
 \begin{minipage}[t!]{0.5\textwidth}
  \centering
     \makeatletter\def\@captype{table}\makeatother\caption{Ablation study on the number $K$ of points selected during the farthest point sampling step in the bottom-up based proposal generation module.}
     \setlength{\tabcolsep}{3.0mm}{
       \begin{tabular}{l|c|c}
\hline
\multicolumn{1}{c|}{K} & Acc@0.25 & Acc@0.5 \\ \hline
256                    & 45.90  & 34.50   \\
128                    & \textbf{49.04}& \textbf{37.80}   \\
64                     & 45.83  & 34.50 \\ \hline
\end{tabular}}
\label{ablation_k}
  \end{minipage}
  \hspace{10pt}
  \begin{minipage}[t!]{0.35\textwidth}
   \centering
        \makeatletter\def\@captype{table}\makeatother\caption{Ablation study on the number $n$ of graph-based information aggregation iterations.}
        \setlength{\tabcolsep}{3.0mm}{
         \begin{tabular}{l|c|c}
\hline
\multicolumn{1}{c|}{n} & Acc@0.25 & Acc@0.5 \\ \hline
0                    & 45.30  & 34.32   \\
1                    &\textbf{49.04} & \textbf{37.80}        \\
2                    & 44.81  & 33.28 \\ \hline
\end{tabular}}
\label{ablation_n}
   \end{minipage}
\end{figure}

\noindent \textbf{Ablation on number $K$ in the farthest point sampling.} Furthermore, we conducted an ablation study on the number of points selected during the farthest point sampling step in the bottom-up based proposal generation module. The experimental results are presented in Table \ref{ablation_k}. From the results shown in Table \ref{ablation_k}, we can conclude that selecting $K$=128 points yields the optimal performance.



\begin{figure*}[t!]
\centering
\vspace{-8pt}
\includegraphics[width=0.9\columnwidth]{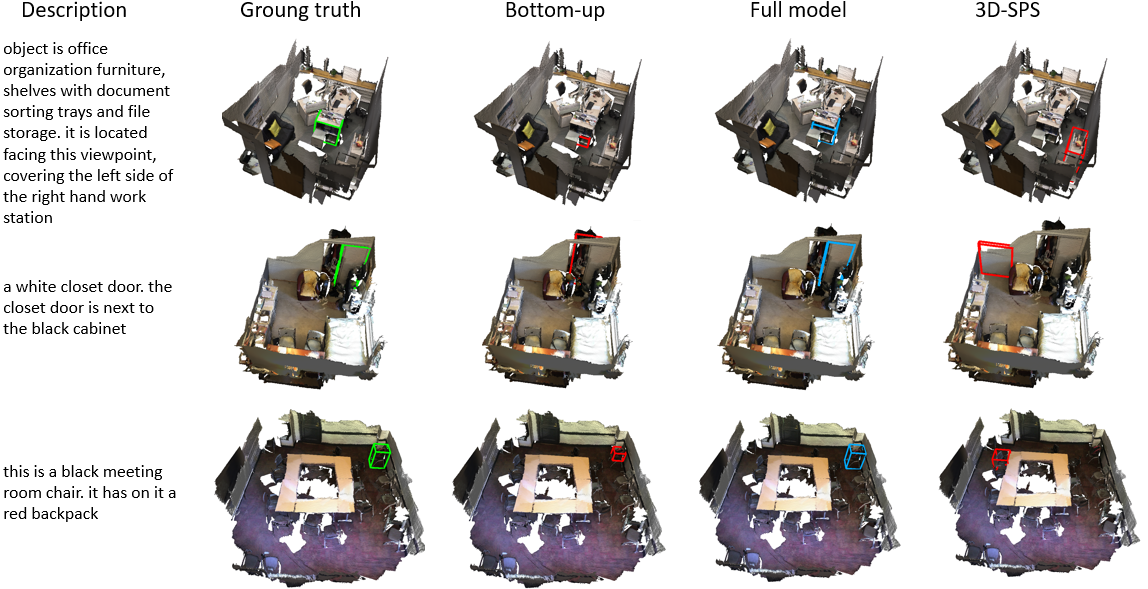}
\vspace{-14pt}
\caption{Visualization of our method. The first column displays the ground truth bounding boxes provided by the ScanRefer dataset. The second and third columns represent the output results of our bottom-up based proposal generation module and the final output of the entire model, respectively. The last column shows the results obtained from the 3D-SPS method.}
    \label{fig:vis}
    \vspace{-14pt}
\end{figure*}

\noindent \textbf{Ablation on graph layers in proposal consolidation module.} We also conducted experiments on the number $n$ of graph-based information aggregation iterations during the top-down stage, and the results are presented in Table \ref{ablation_n}. It can be observed that performing graph-based information aggregation once led to a significant improvement in the results. However, increasing the number of aggregation iterations had a detrimental effect on the results, likely due to factors such as oversmoothing \cite{li2018deeper}, where excessive aggregation on the graph data blurs features.

\textbf{More experiments of ablation studies can be found in our supplementary.}

\subsection{Visualization}
In Figure \ref{fig:vis}, we present a visual comparison of the output results from the two modules of our method, as well as the results obtained from the 3D-SPS \cite{luo20223d} method.
We can see that our bottom-up based proposal generation module, with the aid of query language information, can locate the target objects. However, the quality of the bounding boxes generated in this step often falls short of being optimal. For instance, in the first row, the bounding box only encompasses the lower part of the shelf, and in the third row, the bounding box only covers the upper part of the chair. Nevertheless, through the subsequent top-down based proposal consolidation module and by leveraging information from surrounding proposals, the bounding boxes can be refined to better represent the entire object's position. This consolidation process allows our approach to provide more accurate and complete bounding box predictions, enhancing the overall localization performance. When comparing our method to the 3D-SPS \cite{luo20223d} approach, we can observe that our method excels at precisely localizing the objects described in the query language. \textbf{More visualizations are in our supplementary.}

\section{Conclusion}
In this paper, we have analyzed two categories of methods used in 3D visual grounding: the top-down based method and the bottom-up based method, each with its respective strengths and weaknesses. Our proposed approach integrates the advantages of both methods effectively while alleviating their limitations. Firstly, we utilize a bottom-up based proposal generation module to produce high-quality candidate proposals relevant to the query information. Subsequently, a top-down based consolidation module is employed to further enhance the performance. As a result, our proposed method demonstrates superior performance compared to the state-of-the-art results on the ScanRefer dataset. Furthermore, our approach can serve as a flexible framework, enabling the replacement of both the bottom-up based module and the top-down based module with more advanced methods to achieve even better results in future research.

%
%
%
%
\bibliographystyle{splncs04}
\bibliography{ref}

\begin{thebibliography}{10}
\providecommand{\url}[1]{\texttt{#1}}
\providecommand{\urlprefix}{URL }
\providecommand{\doi}[1]{https://doi.org/#1}

\bibitem{achlioptas2020referit3d}
Achlioptas, P., Abdelreheem, A., Xia, F., Elhoseiny, M., Guibas, L.: Referit3d: Neural listeners for fine-grained 3d object identification in real-world scenes. In: Computer Vision--ECCV 2020: 16th European Conference, Glasgow, UK, August 23--28, 2020, Proceedings, Part I 16. pp. 422--440. Springer (2020)

\bibitem{chen2020scanrefer}
Chen, D.Z., Chang, A.X., Nie{\ss}ner, M.: Scanrefer: 3d object localization in rgb-d scans using natural language. In: Computer Vision--ECCV 2020: 16th European Conference, Glasgow, UK, August 23--28, 2020, Proceedings, Part XX. pp. 202--221. Springer (2020)

\bibitem{chen2021hierarchical}
Chen, S., Fang, J., Zhang, Q., Liu, W., Wang, X.: Hierarchical aggregation for 3d instance segmentation. In: Proceedings of the IEEE/CVF International Conference on Computer Vision. pp. 15467--15476 (2021)

\bibitem{cheng2021back}
Cheng, B., Sheng, L., Shi, S., Yang, M., Xu, D.: Back-tracing representative points for voting-based 3d object detection in point clouds. In: Proceedings of the IEEE/CVF conference on computer vision and pattern recognition. pp. 8963--8972 (2021)

\bibitem{deng2021transvg}
Deng, J., Yang, Z., Chen, T., Zhou, W., Li, H.: Transvg: End-to-end visual grounding with transformers. In: Proceedings of the IEEE/CVF International Conference on Computer Vision. pp. 1769--1779 (2021)

\bibitem{feng2021free}
Feng, M., Li, Z., Li, Q., Zhang, L., Zhang, X., Zhu, G., Zhang, H., Wang, Y., Mian, A.: Free-form description guided 3d visual graph network for object grounding in point cloud. In: Proceedings of the IEEE/CVF International Conference on Computer Vision. pp. 3722--3731 (2021)

\bibitem{he2021transrefer3d}
He, D., Zhao, Y., Luo, J., Hui, T., Huang, S., Zhang, A., Liu, S.: Transrefer3d: Entity-and-relation aware transformer for fine-grained 3d visual grounding. In: Proceedings of the 29th ACM International Conference on Multimedia. pp. 2344--2352 (2021)

\bibitem{hu2022exploring}
Hu, Q., Liu, D., Hu, W.: Exploring the devil in graph spectral domain for 3d point cloud attacks. In: European Conference on Computer Vision. pp. 229--248. Springer (2022)

\bibitem{hu2023density}
Hu, Q., Liu, D., Hu, W.: Density-insensitive unsupervised domain adaption on 3d object detection. In: Proceedings of the IEEE/CVF Conference on Computer Vision and Pattern Recognition. pp. 17556--17566 (2023)

\bibitem{huang2021text}
Huang, P.H., Lee, H.H., Chen, H.T., Liu, T.L.: Text-guided graph neural networks for referring 3d instance segmentation. In: Proceedings of the AAAI Conference on Artificial Intelligence. vol.~35, pp. 1610--1618 (2021)

\bibitem{huang2023dense}
Huang, W., Liu, D., Hu, W.: Dense object grounding in 3d scenes. In: Proceedings of the 31st ACM International Conference on Multimedia. pp. 5017--5026 (2023)

\bibitem{jiang2020pointgroup}
Jiang, L., Zhao, H., Shi, S., Liu, S., Fu, C.W., Jia, J.: Pointgroup: Dual-set point grouping for 3d instance segmentation. In: Proceedings of the IEEE/CVF conference on computer vision and Pattern recognition. pp. 4867--4876 (2020)

\bibitem{li2021referring}
Li, M., Sigal, L.: Referring transformer: A one-step approach to multi-task visual grounding. Advances in neural information processing systems  \textbf{34},  19652--19664 (2021)

\bibitem{li2018deeper}
Li, Q., Han, Z., Wu, X.M.: Deeper insights into graph convolutional networks for semi-supervised learning. In: Proceedings of the AAAI conference on artificial intelligence. vol.~32 (2018)

\bibitem{liu2023exploring}
Liu, D., Fang, X., Hu, W., Zhou, P.: Exploring optical-flow-guided motion and detection-based appearance for temporal sentence grounding. IEEE Transactions on Multimedia  \textbf{25},  8539--8553 (2023)

\bibitem{liu2022imperceptible}
Liu, D., Hu, W.: Imperceptible transfer attack and defense on 3d point cloud classification. IEEE transactions on pattern analysis and machine intelligence  \textbf{45}(4),  4727--4746 (2022)

\bibitem{liu2022learning}
Liu, D., Hu, W.: Learning to focus on the foreground for temporal sentence grounding. In: Proceedings of the 29th International Conference on Computational Linguistics. pp. 5532--5541 (2022)

\bibitem{liu2022rethinking}
Liu, D., Hu, W.: Rethinking graph neural networks for unsupervised video object segmentation. In: BMVC. p.~76 (2022)

\bibitem{liu2022skimming}
Liu, D., Hu, W.: Skimming, locating, then perusing: A human-like framework for natural language video localization. In: Proceedings of the 30th ACM International Conference on Multimedia. pp. 4536--4545 (2022)

\bibitem{liu2024explicitly}
Liu, D., Hu, W.: Explicitly perceiving and preserving the local geometric structures for 3d point cloud attack. In: Proceedings of the AAAI Conference on Artificial Intelligence. vol.~38, pp. 3576--3584 (2024)

\bibitem{liu2023point}
Liu, D., Hu, W., Li, X.: Point cloud attacks in graph spectral domain: When 3d geometry meets graph signal processing. IEEE Transactions on Pattern Analysis and Machine Intelligence  (2023)

\bibitem{liu2023robust}
Liu, D., Hu, W., Li, X.: Robust geometry-dependent attack for 3d point clouds. IEEE Transactions on Multimedia  (2023)

\bibitem{liu2024survey}
Liu, D., Liu, Y., Huang, W., Hu, W.: A survey on text-guided 3d visual grounding: Elements, recent advances, and future directions. arXiv preprint arXiv:2406.05785  (2024)

\bibitem{liu2020saanet}
Liu, D., Ouyang, X., Xu, S., Zhou, P., He, K., Wen, S.: Saanet: Siamese action-units attention network for improving dynamic facial expression recognition. Neurocomputing  \textbf{413},  145--157 (2020)

\bibitem{liu2021context}
Liu, D., Qu, X., Dong, J., Zhou, P., Cheng, Y., Wei, W., Xu, Z., Xie, Y.: Context-aware biaffine localizing network for temporal sentence grounding. In: Proceedings of the IEEE/CVF Conference on Computer Vision and Pattern Recognition. pp. 11235--11244 (2021)

\bibitem{liu2020jointly}
Liu, D., Qu, X., Liu, X.Y., Dong, J., Zhou, P., Xu, Z.: Jointly cross-and self-modal graph attention network for query-based moment localization. In: Proceedings of the 28th ACM International Conference on Multimedia. pp. 4070--4078 (2020)

\bibitem{liu2021spatiotemporal}
Liu, D., Xu, S., Liu, X.Y., Xu, Z., Wei, W., Zhou, P.: Spatiotemporal graph neural network based mask reconstruction for video object segmentation. In: Proceedings of the AAAI Conference on Artificial Intelligence. vol.~35, pp. 2100--2108 (2021)

\bibitem{liu2024a}
Liu, D., Yang, M., Qu, X., Zhou, P., Hu, W., Cheng, Y.: A survey of attacks on large vision-language models: Resources, advances, and future trends. arXiv preprint arXiv:2407.07403  (2024)

\bibitem{liu2021f2net}
Liu, D., Yu, D., Wang, C., Zhou, P.: F2net: Learning to focus on the foreground for unsupervised video object segmentation. In: Proceedings of the AAAI conference on artificial intelligence. vol.~35, pp. 2109--2117 (2021)

\bibitem{liu2021video}
Liu, D., Zhang, H., Zhou, P.: Video-based facial expression recognition using graph convolutional networks. In: 2020 25th International Conference on Pattern Recognition (ICPR). pp. 607--614. IEEE (2021)

\bibitem{liu2021refer}
Liu, H., Lin, A., Han, X., Yang, L., Yu, Y., Cui, S.: Refer-it-in-rgbd: A bottom-up approach for 3d visual grounding in rgbd images. In: Proceedings of the IEEE/CVF Conference on Computer Vision and Pattern Recognition. pp. 6032--6041 (2021)

\bibitem{liu2024cross}
Liu, Y., Liu, D., Guo, Z., Hu, W.: Cross-task knowledge transfer for semi-supervised joint 3d grounding and captioning. In: Proceedings of the 32st ACM International Conference on Multimedia. {ACM} (2024)

\bibitem{liu2021group}
Liu, Z., Zhang, Z., Cao, Y., Hu, H., Tong, X.: Group-free 3d object detection via transformers. In: Proceedings of the IEEE/CVF International Conference on Computer Vision. pp. 2949--2958 (2021)

\bibitem{loshchilov2018fixing}
Loshchilov, I., Hutter, F.: Fixing weight decay regularization in adam  (2018)

\bibitem{luo20223d}
Luo, J., Fu, J., Kong, X., Gao, C., Ren, H., Shen, H., Xia, H., Liu, S.: 3d-sps: Single-stage 3d visual grounding via referred point progressive selection. In: Proceedings of the IEEE/CVF Conference on Computer Vision and Pattern Recognition. pp. 16454--16463 (2022)

\bibitem{misra2021end}
Misra, I., Girdhar, R., Joulin, A.: An end-to-end transformer model for 3d object detection. In: Proceedings of the IEEE/CVF International Conference on Computer Vision. pp. 2906--2917 (2021)

\bibitem{qi2019deep}
Qi, C.R., Litany, O., He, K., Guibas, L.J.: Deep hough voting for 3d object detection in point clouds. In: proceedings of the IEEE/CVF International Conference on Computer Vision. pp. 9277--9286 (2019)

\bibitem{qi2017pointnet}
Qi, C.R., Su, H., Mo, K., Guibas, L.J.: Pointnet: Deep learning on point sets for 3d classification and segmentation. In: Proceedings of the IEEE conference on computer vision and pattern recognition. pp. 652--660 (2017)

\bibitem{qi2017pointnet++}
Qi, C.R., Yi, L., Su, H., Guibas, L.J.: Pointnet++: Deep hierarchical feature learning on point sets in a metric space. Advances in neural information processing systems  \textbf{30} (2017)

\bibitem{radford2021learning}
Radford, A., Kim, J.W., Hallacy, C., Ramesh, A., Goh, G., Agarwal, S., Sastry, G., Askell, A., Mishkin, P., Clark, J., et~al.: Learning transferable visual models from natural language supervision. In: International conference on machine learning. pp. 8748--8763. PMLR (2021)

\bibitem{radosavovic2020designing}
Radosavovic, I., Kosaraju, R.P., Girshick, R., He, K., Doll{\'a}r, P.: Designing network design spaces. In: Proceedings of the IEEE/CVF conference on computer vision and pattern recognition. pp. 10428--10436 (2020)

\bibitem{sadhu2019zero}
Sadhu, A., Chen, K., Nevatia, R.: Zero-shot grounding of objects from natural language queries. In: Proceedings of the IEEE/CVF International Conference on Computer Vision. pp. 4694--4703 (2019)

\bibitem{tao20233dhacker}
Tao, Y., Liu, D., Zhou, P., Xie, Y., Du, W., Hu, W.: 3dhacker: Spectrum-based decision boundary generation for hard-label 3d point cloud attack. In: Proceedings of the IEEE/CVF International Conference on Computer Vision. pp. 14340--14350 (2023)

\bibitem{vaswani2017attention}
Vaswani, A., Shazeer, N., Parmar, N., Uszkoreit, J., Jones, L., Gomez, A.N., Kaiser, {\L}., Polosukhin, I.: Attention is all you need. Advances in neural information processing systems  \textbf{30} (2017)

\bibitem{vu2022softgroup}
Vu, T., Kim, K., Luu, T.M., Nguyen, T., Yoo, C.D.: Softgroup for 3d instance segmentation on point clouds. In: Proceedings of the IEEE/CVF Conference on Computer Vision and Pattern Recognition. pp. 2708--2717 (2022)

\bibitem{wang2018learning}
Wang, L., Li, Y., Huang, J., Lazebnik, S.: Learning two-branch neural networks for image-text matching tasks. IEEE Transactions on Pattern Analysis and Machine Intelligence  \textbf{41}(2),  394--407 (2018)

\bibitem{wang2019neighbourhood}
Wang, P., Wu, Q., Cao, J., Shen, C., Gao, L., Hengel, A.v.d.: Neighbourhood watch: Referring expression comprehension via language-guided graph attention networks. In: Proceedings of the IEEE/CVF Conference on Computer Vision and Pattern Recognition. pp. 1960--1968 (2019)

\bibitem{wang2019dynamic}
Wang, Y., Sun, Y., Liu, Z., Sarma, S.E., Bronstein, M.M., Solomon, J.M.: Dynamic graph cnn for learning on point clouds. ACM Transactions on Graphics (tog)  \textbf{38}(5),  1--12 (2019)

\bibitem{yang2022tubedetr}
Yang, A., Miech, A., Sivic, J., Laptev, I., Schmid, C.: Tubedetr: Spatio-temporal video grounding with transformers. In: Proceedings of the IEEE/CVF Conference on Computer Vision and Pattern Recognition. pp. 16442--16453 (2022)

\bibitem{yang2019cross}
Yang, S., Li, G., Yu, Y.: Cross-modal relationship inference for grounding referring expressions. In: Proceedings of the IEEE/CVF conference on computer vision and pattern recognition. pp. 4145--4154 (2019)

\bibitem{yang2019fast}
Yang, Z., Gong, B., Wang, L., Huang, W., Yu, D., Luo, J.: A fast and accurate one-stage approach to visual grounding. In: Proceedings of the IEEE/CVF International Conference on Computer Vision. pp. 4683--4693 (2019)

\bibitem{yang2021sat}
Yang, Z., Zhang, S., Wang, L., Luo, J.: Sat: 2d semantics assisted training for 3d visual grounding. In: Proceedings of the IEEE/CVF International Conference on Computer Vision. pp. 1856--1866 (2021)

\bibitem{yuan2021instancerefer}
Yuan, Z., Yan, X., Liao, Y., Zhang, R., Wang, S., Li, Z., Cui, S.: Instancerefer: Cooperative holistic understanding for visual grounding on point clouds through instance multi-level contextual referring. In: Proceedings of the IEEE/CVF International Conference on Computer Vision. pp. 1791--1800 (2021)

\bibitem{zhao20213dvg}
Zhao, L., Cai, D., Sheng, L., Xu, D.: 3dvg-transformer: Relation modeling for visual grounding on point clouds. In: Proceedings of the IEEE/CVF International Conference on Computer Vision. pp. 2928--2937 (2021)

\end{thebibliography}




\end{document}